\documentclass[english,conference,a4paper]{IEEEtran}
\usepackage[cp1250]{inputenc}
\usepackage{array}
\usepackage{units}
\usepackage{textcomp}
\usepackage{multirow}
\usepackage{amsmath}
\usepackage{amssymb}
\usepackage{graphicx}
\usepackage{epstopdf}

\usepackage[colorinlistoftodos, shadow]{todonotes/todonotes}

\makeatletter

\providecommand{\tabularnewline}{\\}

\makeatother

\usepackage{babel}
\begin{document}

\title{Towards Deep Compositional Networks}


\author{
\IEEEauthorblockN{Domen Tabernik\IEEEauthorrefmark{1}\IEEEauthorrefmark{2}, Matej Kristan\IEEEauthorrefmark{2},
Jeremy L. Wyatt\IEEEauthorrefmark{1}, Ale\v{s} Leonardis\IEEEauthorrefmark{1}\IEEEauthorrefmark{2}}
\IEEEauthorblockA{\IEEEauthorrefmark{1}CN-CR Centre, School of Computer Science, \\University
of Birmingham, Birmingham, UK}
\IEEEauthorblockA{\IEEEauthorrefmark{2}Faculty of Computer and Information Science,\\
University of Ljubljana, Ljubljana, Slovenia\\
\{domen.tabernik,matej.kristan,ales.leonardis\}@fri.uni-lj.si\\
\{a.leonardis,jlw\}@cs.bham.ac.uk}
}

\maketitle

\author{\IEEEauthorblockN{Domen Tabernik}
\IEEEauthorblockA{CN-CR Centre,\\School of Computer Science,\\University of Birmingham\\and\\}
\IEEEauthorblockA{Faculty of Computer and\\Information Science,\\University of Ljubljana}\and
\IEEEauthorblockN{Matej Kristan}
\IEEEauthorblockA{Faculty of Computer and\\Information Science,\\University of Ljubljana}\and
\IEEEauthorblockN{Jeremy Wyatt}
\IEEEauthorblockA{CN-CR Centre,\\School of Computer Science,\\University of Birmingham}
\and
\IEEEauthorblockN{Ale\v{s} Leonardis}
\IEEEauthorblockA{CN-CR Centre,\\School of Computer Science,\\University of Birmingham\\and\\}
\IEEEauthorblockA{Faculty of Computer and\\Information Science,\\University of Ljubljana}}
\begin{abstract}
Hierarchical feature learning based on convolutional neural networks (CNN) has recently shown significant potential in various computer vision tasks. While allowing high-quality discriminative feature learning, the downside of CNNs is the lack of explicit structure in features, which often leads to overfitting, absence of reconstruction from partial observations and limited generative abilities. Explicit structure is inherent in hierarchical compositional models, however, these lack the ability to optimize a well-defined cost function. We propose a novel analytic model of a basic unit in a layered hierarchical model with both explicit compositional structure and a well-defined discriminative cost function. Our experiments on two datasets show that the proposed compositional model performs on a par with standard CNNs on discriminative tasks, while, due to explicit modeling of the structure in the feature units, affording a straight-forward visualization of parts and faster inference due to separability of the units.

\end{abstract}

\section{Introduction}

Over the last ten years, computer-vision-based object perception has continuously been moving away from using hand-crafted features like HOG~\cite{Dalal05histogramsof} and SIFT~\cite{Lowe1999}, and significant efforts have been made on feature learning. 
An important characteristics of state-of-the-art feature learning methods~\cite{Krizhevsky2012,Lecun2006,Leonardis2011,Zhu2010} can be found in the graduated complexity of the features as layers are added to the network, thus forming a hierarchy.

\begin{figure}

\includegraphics[width=1\columnwidth]{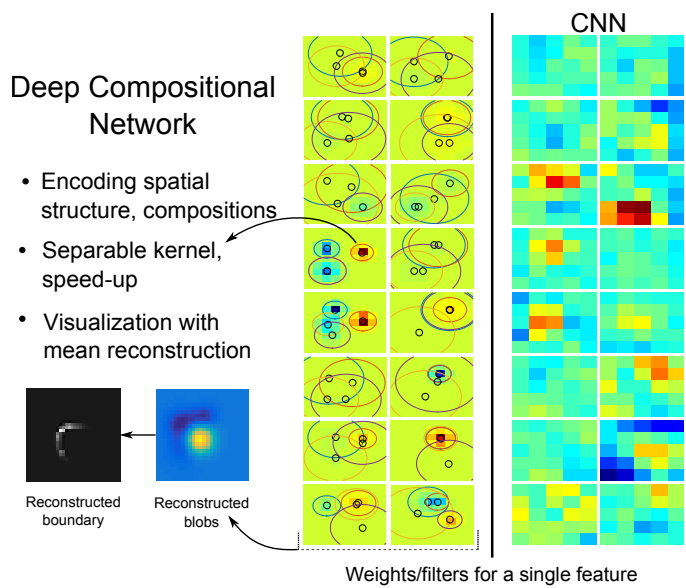}\vspace{-5pt}\caption{\label{fig:intro-img} Units learned by our deep compositional network (left) and equivalent units learned by the convolutional neural network (right).}
\vspace{-12pt}
\end{figure}

Two paradigms have emerged in the design of hierarchical models, that differ in the definition of a unit learned at each layer. The first paradigm is that of compositional models~\cite{Fidler07towardsscalable,SiPAMI2013}. We refer to models as being compositional when its features at a layer are modeled as an explicit combination of features in the lower layer. These models allow fast inference via inverted indexing, inherently produce region proposals, offer straight-forward reconstruction from partial observations and visualization of features. However, the learning cost function is weakly defined and is usually performed via co-occurrence learning~\cite{Fidler07towardsscalable,SiPAMI2013}. While attempts have been made to discriminatively re-interpret the learned reconstructive parts~\cite{Kristan2013,Tabernik2015a}, learning discriminative parts directly remains an open issue. On the other hand, the paradigm of convolutional models has recently gained significant momentum in feature learning. These models define the feature units at each layer as filters, which afford learning by back-propagation. In particular, convolutional neural networks (CNNs)~\cite{Lecun2006,Krizhevsky2012} have emerged as a highly successful representative of this class of hierarchical models. A longstanding criticism of CNNs is the lack of precise spatial relationships between high-level parts, a reason advocated to move towards viewpoint-invariant capsule-like systems~\cite{Hinton2011}. The lack of structure in basic units in CNNs also hinders robust handling of occlusions and missing parts~\cite{Yuille2013a,Tabernik2015a}, and prohibits straight-forward visualization and understanding of networks. Approximate visualization techniques have been developed~\cite{Zeiler2013,Mahendran2015, Mahendran2015a}, but further attempts at understanding CNNs uncovered unintuitive behavior when small input perturbations were applied~\cite{Szegedy2013}. Recent efforts have also been made toward approximation of the learned filters to reduce the computational complexity of the learned CNNs~\cite{Jaderberg2014}, an issue typically addressed by brute force CPU/GPU power increase.

In this paper we propose a novel form of the unit in a deep hierarchical model with an explicit compositional structure (Fig.~\ref{fig:intro-img}). While stacked weights in deep networks can be considered as compositions, they lack an explicit structure that could further expose and leverage compositional properties. Our proposed unit exposes an explicit structure of compositions as a parametric model over spatial clustering of responses from the lower layer, and can directly be embedded into the learning framework used in CNNs. This allows learning of hierarchical compositional models with a well-defined, potentially discriminative, cost function similar to that used in convolutional neural networks, and retains the benefits of compositional models, such as a precise encoding of the spatial relationship between parts. We derive the necessary equations for back-propagation and propose a compositional model trained by a discriminative cost function, which is the major contribution of our paper. We experimentally evaluate the proposed model on CIFAR-10~\cite{Krizhevsky2009} and PaCMan~\cite{PacManDB} datasets and show that our model achieves comparable performance to a standard CNN, while allowing simple compositional mean-reconstruction of parts. Since our units are separable by design, we demonstrate a significant speedup in inference compared to CNNs. 

The paper is structured as follows: Section~\ref{sec:gaussian-parametrization}
describes our model, Section~\ref{sec:evaluation} provides
experimental evaluation, in
Section~\ref{sec:visualization-and-speed-up} visualization and speed-up in inference are presented, and conclusions are drawn in Section~\ref{sec:conclusion}.

\section{Deep compositional network\label{sec:gaussian-parametrization}}

We first provide notation for deep convolutional neural networks and
then derive our compositional network. A neuron activation in a deep neural network is modeled with a linear function wrapped inside a specific non-linearity: \mbox{
$y_{i}=f(\sum\nolimits_s w_{s}\cdot x_{s}+b_{s})$},
with output activation $y$, input activations $x$, bias $b$ and
weights $w$ determining a linear combination of input activations further
modified by a non-linear function $f(\cdot)$. In the image domain the neuron activations are organized in a 3-dimensional
matrix $N\times M\times S$, where two dimensions, $N\times M$, represent the
image or feature plane, and the third dimension, $S$ represents channels. 
With the introduction of weight-sharing along the 2D feature
plane, the convolutional neural network models the neuron's activation function $Y$ as:
\begin{equation}
Y_{i}=f(\sum\nolimits_s W_{s}\ast X_{s}+b_{s}),\label{eq:cnn-conv}
\end{equation}
where $\ast$ is a convolution of $X_{s}$, the
$N\times M$ activation map from the $s$-th channel, with $W_{s}$, the $K_w\times K_h$
weights for the $s$-th channel. The  $W_{s}$ are basic units in the CNNs that take the form of convolution filters and have to be learned from the data. The convolutions give an $N-K_w\times M-K_h$
output map for each channel. Element-wise summation over the channels
represents the final activation map $Y_{i}$ after non-linearity $f(\cdot)$
is applied. Typically, several activation maps $Y_{i}$ are
created, and the network is organized in layers, such that $X_{s}$ in the
$l$-th layer is the output activation $Y_{i}$ from $l-1$ layer.

We propose a new basic unit that explicitly models the composition of a feature. We define it as a weighted Gaussian component:
 \begin{equation}
 \tilde{W}_{k}=\tilde{w}_{k}G(\theta_k),\label{eq:new-unit}
 \end{equation} 
with weight $\tilde{w}_{k}$ and Gaussian parameters $\theta_k=[\vec{\mu}_{k},\sigma_{k}]$, containing mean $\vec{\mu}_{k}$ and variance $\sigma_{k}^{2}$. Multiple $\tilde{W}_{k}$ can be applied to the same input channel $s$, but we  omit  this in the notation in the interest of clarity. 
Units $\tilde{W}_{k}$ take the form of convolution filters and have
 $K_w\times K_h$ elements. We therefore define $G(\cdot)$ as a two dimensional matrix
of the same size with 2-dimensional index $\vec{x}$ over $K_w\times K_h$ elements:
 \begin{equation}
G(\vec{x},\theta)=\frac{1}{\sqrt{2\pi\sigma^{2}}}exp(-\frac{\left\Vert \vec{x}-\vec{\mu}\right\Vert^{2}}{2\sigma^{2}}).
\end{equation}
We use two dimensional means but single dimensional variance for simplification.
Commonly used unit sizes are $3\times3$, $5\times5$ or $9\times9$;
however, such small sizes can lead to significant discretization errors
in $G(\vec{x},\theta_k)$. We avoid this by replacing the normalization
factor computed in continuous space with one computed in the discretized
space, leading to our final distribution function $G(\vec{x},\theta)$:

 \begin{equation}
G(\vec{x},\theta)=\frac{1}{N(\theta)}g(\vec{x},\theta),
\end{equation}
where $g(\vec{x},\theta)$ is a non-normalized Gaussian distribution
and $N(\theta)$ is a sum over this non-normalized Gaussian distribution computed for a filter of size $K_w\times K_h$: 
 \begin{equation}
\begin{array}{c}
N(\theta)=\sum\nolimits_{\vec{x}}g(\vec{x},\theta),\,\,
g(\vec{x},\theta)=exp(-\frac{\left\Vert \vec{x}-\vec{\mu}\right\Vert^{2}}{2\sigma^{2}}).
\end{array}
\end{equation}

A new, compositional unit can be embedded into a CNN by grouping multiple instances applied to the same input channel $s$ and deriving the basic CNN unit from Eq.~(\ref{eq:cnn-conv}):
 \begin{equation}
 W_{s}=\sum\nolimits_k \tilde{W}_{k}=\sum\nolimits_k\tilde{w}_{k}G(\theta_k).\label{eq:weight-parametrization}
 \end{equation}
This proposed model for each $W_{s}$ unit is similar to a standard Gaussian mixture model, but we do not enforce $\sum\tilde{w}=1$ and
component weights can take any value, $\tilde{w}\in[-\infty,\infty]$.
Having negative weights is important to approximate edges as differences
between neighboring components, whereas positive components can be interpreted
as requirements for a presence of a feature and negative components
as requirements for an absence of a feature. In principle, this could
be satisfied by normalizing the sum of absolute weights,
$\sum\left|\tilde{w}\right|=1$, but this  would significantly complicate the gradient computation without any performance gain.

Learning an individual unit $\tilde{W}_{k}$ consists of learning its parameters for the Gaussian distribution, mean $\vec{\mu}_{k}$ and variance $\sigma_{k}^{2}$, together with the weight $\tilde{w}_{k}$.
The number of units, i.e., the number of Gaussian components per input channel, can be considered as a
hyper-parameter. Learning can be performed in the same way as in convolutional
networks via gradient descent. Parameters
are optimized by computing the gradients w.r.t. the cost function $C(y,\bar{y})$,
which leads to three different types of gradients. By applying the chain
rule we can define the gradient for the component weight $\nicefrac{\partial C}{\partial\tilde{w}_{k}}$
as a dot-product of back-propagated error and the input feature $X$$_{s}$
convolved with the $k$-th Gaussian component:

\begin{figure*}
\includegraphics[width=1\textwidth]{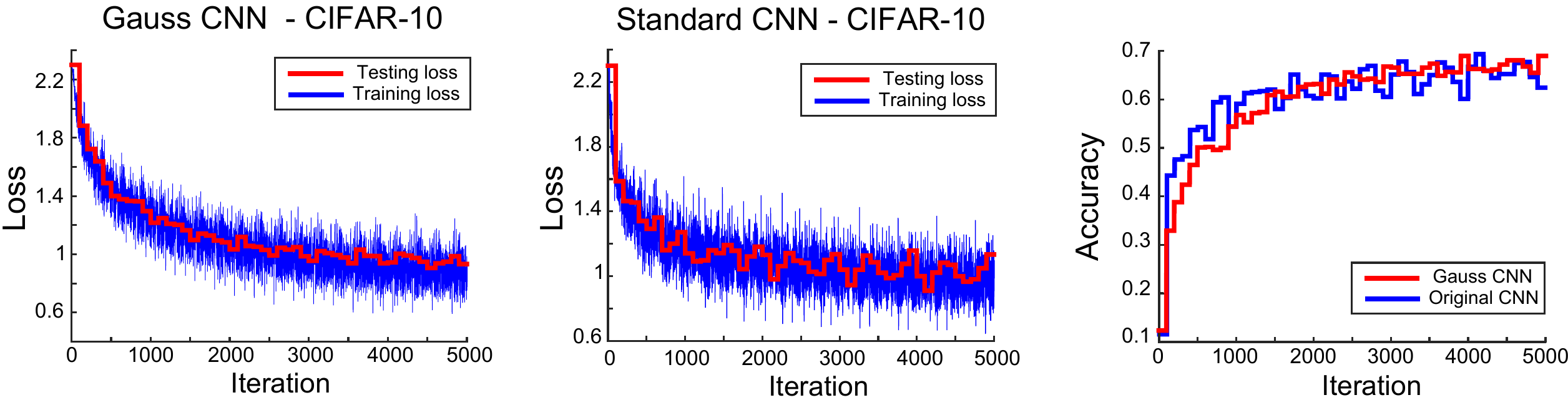}\vspace{-7pt}\caption{\label{fig:results-cifar10}Classification results on CIFAR-10 database.}
\vspace{-10pt}
\end{figure*}

\begin{equation}
\frac{\partial C}{\partial\tilde{w}_{k}}=\underset{n,m}{\sum}\frac{\partial C}{\partial Z}\cdot\frac{\partial Z}{\partial\tilde{w}_{k}}=\underset{n,m}{\sum}\frac{\partial C}{\partial Z}\cdot\underset{\vec{x}}{\sum}X_{s}\ast G(\vec{x},\theta),\label{eq:gradient-c-wrt-weight}
\end{equation}
where $Z=\sum_s W_{s}\ast X_{s}+b_{s}$ and $\nicefrac{\partial C}{\partial Z}$
is back-propagated error. Note, that only the $s$-th channel of input
features are used since the weight component $\tilde{w}_{k}$ appears
only in $W_{s}$. The back-propagated error for layer $l$ is computed the same as in a
standard convolutional network:

\begin{equation}
\frac{\partial C}{\partial Z_{s}^{l}}=\frac{\partial C}{\partial Z_{s}^{l+1}}\ast rot(W_{s}),
\end{equation}
where the back-propagated error from the higher layer $l+1$ is convolved with
the 180\textdegree{} rotated unit (a weight filter) $rot(W_{s})$ which can be computed from Eq.~\ref{eq:weight-parametrization}.
We can similarly apply the chain rule to obtain the gradient for the mean and the
variance:

\begin{equation}
\frac{\partial C}{\partial\mu_{k}}=\underset{n,m}{\sum}\frac{\partial C}{\partial Z}\cdot \underset{\vec{x}}{\sum} X_{s}\ast\frac{\partial G(\vec{x},\theta_k)}{\partial\mu_{k}},\label{eq:gradient-c-wrt-mean}
\end{equation}

\begin{equation}
\frac{\partial C}{\partial\sigma_{k}}=\underset{n,m}{\sum}\frac{\partial C}{\partial Z}\cdot \underset{\vec{x}}{\sum} X_{s}\ast\frac{\partial G(\vec{x},\theta_k)}{\partial\sigma_{k}},\label{eq:gradient-c-wrt-variance}
\end{equation}
where the derivatives of the Gaussian are: 

{\small 
 \begin{equation}
\frac{\partial G(\vec{x},\theta_k)}{\partial\mu_{k}}=\tilde{w}_{k}\frac{N(\theta_k)\cdot\frac{g(\vec{x},\theta_k)}{\partial\mu_{k}}-g(\vec{x},\theta_k)\cdot\frac{\partial N(\theta_k)}{\partial\mu_{k}}}{\left[N(\theta_k)\right]^{2}},
\end{equation}
}{\small \par}

{\small 
 \begin{equation}
\frac{\partial G(\vec{x},\theta_k)}{\partial\sigma_{k}}=\tilde{w}_{k}\frac{N(\theta_k)\cdot\frac{g(\vec{x},\theta_k)}{\partial\sigma_{k}}-g(\vec{x},\theta_k)\cdot\frac{\partial N(\theta_k)}{\partial\sigma_{k}}}{\left[N(\theta_k)\right]^{2}}.
\end{equation}
}

\section{Classification performance\label{sec:evaluation}}

This section analyzes the discriminative properties of our deep compositional network. The method is evaluated on two classification datasets, CIFAR-10~\cite{Krizhevsky2009} and PaCMan~\cite{PacManDB}.

The evaluation on both datasets is performed with a network containing
three layers. The first two layers are convolutional/compositional and the third one
is fully-connected. We use soft-max with multinominal logistic loss
as the cost function. Either three-channel RGB (CIFAR-10) or a single-channel
gray-scale image (PaCMan) is used as input data with zero-mean normalization.
The data is not normalized to unit variance, since between each layer
we use ReLU non-linearity which is less sensitive to data variance.
Note that we use slightly bigger filters (basic units) in our network, but
use fewer components to approximately match the number of parameters
with the standard CNN model. We also restrict components' positions
and standard deviation to ensure derivative of the resulting filters would not have non-zero values outside of the valid window. This also
prevents collapsing to a single point and stalling. Positions
are restricted to at least $1.5$ pixels away from the borders of the valid window,
and standard deviations are restricted to $\sigma_{s}>0.5$. We apply the AdaDelta~\cite{Zeiler2012} with momentum of $0.8$
and no weight-decay to achieve proper behavior in gradient descent.

\begin{table}
\caption{\label{tab:conf-cifar10}Configuration table for neural network used
on CIFAR-10 database.}
\vspace{-7pt}
\centering{}%
\begin{tabular}{cccccc}
\hline 
\multicolumn{1}{c}{} &  &  & \multicolumn{2}{c}{Deep Compositional } & standard \tabularnewline
\multicolumn{1}{c}{} &  &  & \multicolumn{2}{c}{Network (our)} & CNN\tabularnewline
\hline 
\multicolumn{1}{c}{} & num  & \multirow{2}{*}{stride} & unit/filter  & num & unit/filter\tabularnewline
\multicolumn{1}{c}{} & features &  & size & components & size\tabularnewline
\hline \hline 
conv1 & 32 & 1 & $7\times7$ & $2\times2$ & $5\times5$\tabularnewline
relu1 & \multirow{2}{*}{/} & 1 & / & \multirow{2}{*}{/} & \multirow{2}{*}{/}\tabularnewline
pool1 &  & 1 & $3\times3$ &  & \tabularnewline
\hline 
conv2 & 32 & 1 & $9\times9$ & $3\times3$ & $5\times5$\tabularnewline
relu2 & \multirow{2}{*}{/} & 1 & / & \multirow{2}{*}{/} & \multirow{2}{*}{/}\tabularnewline
pool2 &  & 2 & $3\times3$ &  & \tabularnewline
\hline 
ip1 & 10 & / & $15\times15$ & / & $15\times15$\tabularnewline
\hline 
\end{tabular}
\vspace{-10pt}
\end{table}

\subsection{Classification on CIFAR-10}
The CIFAR-10~\cite{Krizhevsky2009} dataset consists of 60.000 images split into 50.000 training
and 10.000 testing images. We perform training with a mini-batch size
of 100 images per iteration and run learning for 5000 iterations.
A detailed configuration of the network used is shown in Table~\ref{tab:conf-cifar10}.
Comparing performance of both models, shown in Fig.~\ref{fig:results-cifar10},
we can see the same accuracy with both achieving slightly less than
70\% on testing set. The left-most two graphs reveal similar learning rates with standard CNN learning slightly faster, but they both
converge to a similar loss in the end. 

\subsection{Classification on PaCMan database}

\begin{figure*}
\includegraphics[width=1\textwidth]{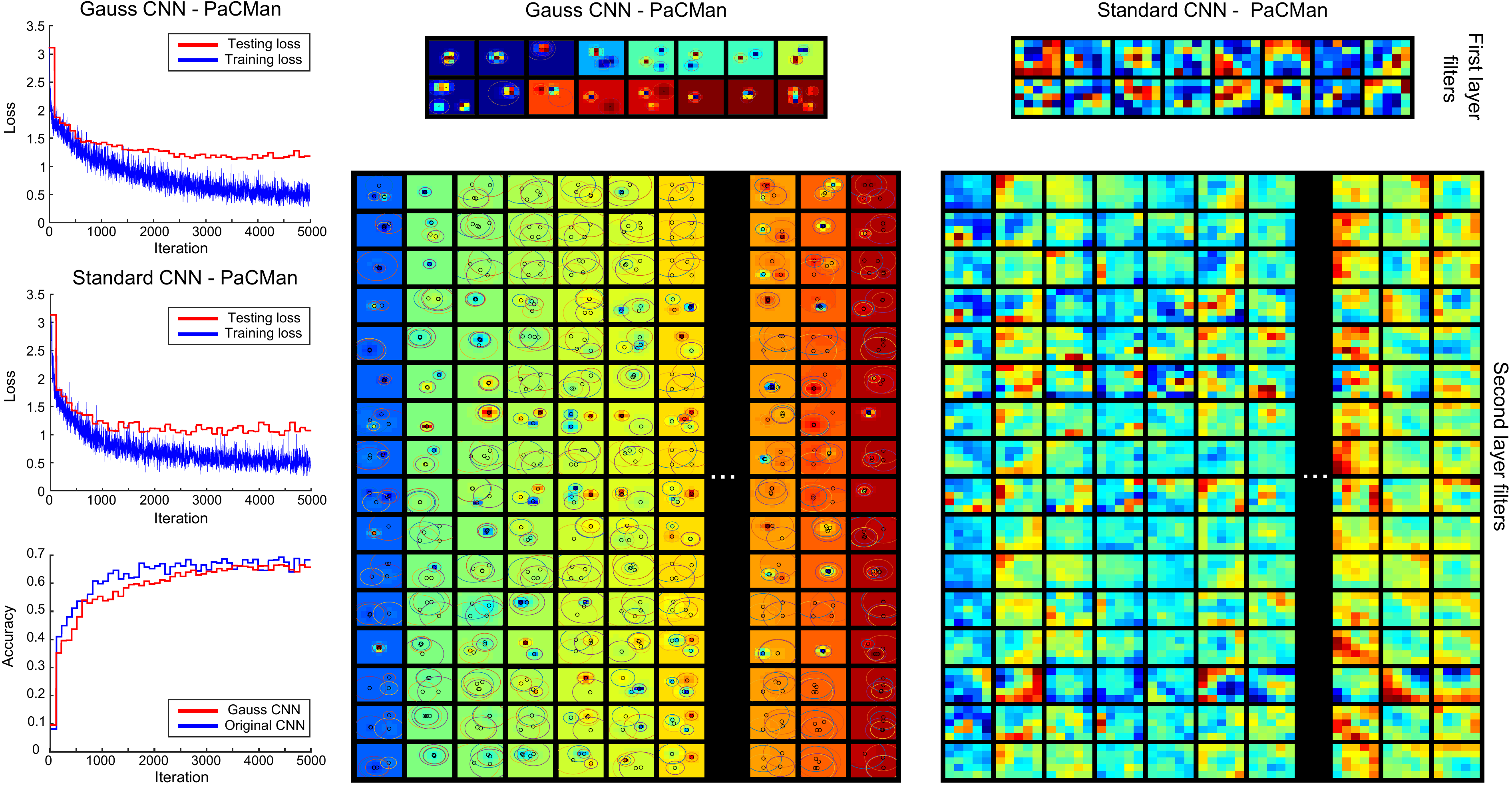}\vspace{-10pt}\caption{\label{fig:results-pacman}Evaluation results on PaCMan database.
Note, only a subset of first-layer and second-layer features are shown.
Each first-layer feature consists of one filter, and each second-layer
feature consists of 16 filters. Visualized weights trained with Gaussian modeling
have component's means depicted as small circles, and variance as big circle. Note, background color within each feature (i.e. individual patches in first layer and patches within column for the second layer) is considered as zero value. Background colors differ between different features due to different max and min range of weights.}
\vspace{-12pt}
\end{figure*}

\begin{table}
\caption{\label{tab:conf-pacman}Configuration table for neural network used
on PaCMan database.}
\vspace{-7pt}
\centering{}%
\begin{tabular}{cccccc}
\hline
\multicolumn{1}{c}{} &  &  & \multicolumn{2}{c}{Deep Compositional } & standard\tabularnewline
\multicolumn{1}{c}{} &  &  & \multicolumn{2}{c}{Network (our)} & CNN\tabularnewline
\hline
\multicolumn{1}{c}{} & num & stride & unit/filter  & num & unit/filter \tabularnewline
\multicolumn{1}{c}{} & features &  & size & components & size\tabularnewline
\hline \hline
conv1 & 16 & 1 & $9\times9$ & $2\times2$ & $5\times5$\tabularnewline
relu1 & \multirow{2}{*}{/} & 1 & / & \multirow{2}{*}{/} & \multirow{2}{*}{/}\tabularnewline
pool1 &  & 1 & $3\times3$ &  & \tabularnewline
\hline 
conv2 & 16 & 1 & $9\times9$ & $2\times2$ & $5\times5$\tabularnewline
relu2 & \multirow{2}{*}{/} & 1 & / & \multirow{2}{*}{/} & \multirow{2}{*}{/}\tabularnewline
pool2 &  & 2 & $3\times3$ &  & \tabularnewline
\hline 
ip1 & 20 & / & $15\times15$ & / & $15\times15$\tabularnewline
\hline
\end{tabular}
\vspace{-10pt}
\end{table}

The proposed compositional network was also evaluated on the PaCMan dataset~\cite{PacManDB}. This dataset contains gray-scale and depth images generated from 3D models of 20 categories
of various kitchen objects, with each category containing 20 different
instances of objects. Each object is captured at dense, regular viewpoint
intervals, but we use only 28 different viewpoints, summing to a total
of around 50.000 images. We use only gray-scale images. The dataset was
split into approximately 25.000 samples for training and 25.000 for testing.
We ensure that all viewpoints of the same object are in the same split
and each category has proportionally the same number of objects in
testing and training. The input images are resized to $128\times96$ to
fit the network into the GPU. The network configuration as shown in Table~\ref{tab:conf-pacman} is slightly modified
to accommodate higher resolution images.

Fig.~\ref{fig:results-pacman} confirms that the proposed compositional network achieves discriminative performance on a par with the CNN -- both models attain an accuracy of 64-67\%. 

We also visualize CNN filters and units in our network for features on the first and the second layer in Fig.~\ref{fig:results-pacman}. The units in our network applied to the same input channel are visualized in a single filter based on Eq.~(\ref{eq:weight-parametrization}), while individual component's means are plotted as small circles, and variances as large circles.
Filters in the standard CNN have a certain structure but they are still
noisy and incoherent. It would also be difficult to capture this structure
without human interpretation. On the other hand, our unit with Gaussian models explicitly captures the spatial structure as can be seen in the first-layer filters. Many components converge to the same location
and the configuration of components directly points to different
edge or blob detectors. On the second layer only
a small set of components have high weights, indicating
that most are irrelevant for the final classification, offering further simplifications of the network. Compared to the filters from the standard CNN they are more compact. With Gaussian modeling a spatial position of a sub-feature is much clearer and easily determined, and, as we show in the next section, can be further utilized.

\section{Utilizing compositional representation\label{sec:visualization-and-speed-up}}

In this section we demonstrate two advantages of having a rich, compositional representation. We propose a novel visualization of deep networks using mean reconstruction of compositions and demonstrate the inference speed-up due to inherent kernel separability in compositions.

\subsection{Visualization by mean reconstruction }

Feature visualization in standard deep networks is difficult due
to lack of structure. Visualization is thus usually performed indirectly by
deconvolution~\cite{Zeiler2013} or optimization of
input pixels to maximize the output activations~\cite{Mahendran2015,Mahendran2015a}.
Such visualization techniques are applicable to our model as well,
but having explicit compositions enables exploration of more straight-forward visualization techniques. 

A feature can be visualized 
by finding an image that produces a maximum output response for its unit. Based on Eq.~(\ref{eq:cnn-conv}) a maximum
response is obtained when all sub-features match to specific patterns defined in weights $W_s$.
In our model each $W_s$ consists of individual compositions $\tilde{W}_k$, thus the premise can further be extended to having a maximum response when individual sub-features
are present at specific position, i.e., at a mean of a Gaussian in our
case. We propose to visualize a single feature by recursively
projecting compositions top-down to image pixels by following (indexing)
the corresponding means in network units. We term this process as mean reconstruction, similar to visualization techniques for hierarchical compositions~\cite{Fidler2009}. 

\begin{figure}
\includegraphics[width=1\columnwidth]{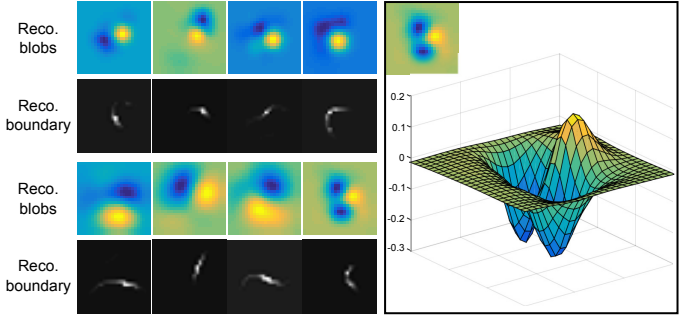}\caption{\label{fig:mean-reconstruction-pacman}Mean reconstruction of 8 features from the second layer of PaCMan dataset. The first and the third rows depict
compositions and their variances back-projected to pixel values. They
can be considered as detectors for specific blob configurations, where yellow values request high positive values, and dark blue, low negative values. 3D visualization of the last feature is also on the left. The second and the fourth rows then show corresponding edge configurations between positive (yellow) and negative (blue) components.}
\vspace{-12pt}
\end{figure}

During each step of back-projection, several properties of the compositions
need to be correctly accounted for: (a) expected uncertainty
of the position of a sub-feature (variance), (b) the importance of a sub-feature (weight) and (c) requested
presence or absence of a sub-feature (weight sign). The uncertainty is defined by a 
Gaussian variance and grows with each step of back-projection. We account
for this by summing the variances along each step to arrive at the final
uncertainty at the pixel level. The importance of a sub-feature is accounted
for by multiplying the magnitude of weight at each back-projected
step. We consider only sub-feature components with a positive
weight, i.e., ones that request the presence of a sub-feature. Sub-feature components with negative weights are ignored. This is applied to all layers, except the first one, since negative values of features are truncated by the ReLU layer on all layers except for the first one. We therefore consider the first layer as a special case and use compositions with negative weights as well.

Positive and negative weights in the first layer in most cases define
edges. But edges are defined indirectly since means in Gaussians
define positions of blobs, i.e., regions with low intensity or high
intensity values. Edges can be inferred from neighboring blobs with
opposing signs and are free to occur anywhere between blobs, either
as a smooth transition or a sharp edge. Consequently, after all
compositions are back-projected from the top to the bottom we are
left with two sets of Gaussian distributions, ones with the positive sign
and ones with the negative sign. They are visualized in the second and the fourth row of Fig.~\ref{fig:mean-reconstruction-pacman}. This visualization
is performed by summing over all positive and negative distributions
for each pixel. We refer to a map obtained this way as a reconstructed
blobs or distribution map.

We can further visualize Gaussian compositions by their boundaries that separate positive and negative components. We achieve this by translating the problem to a graph-cut problem,
where pixels are represented as a graphical model connected to neighbors 
to either a sink or a source. We use a sum over all positive distributions
for one pixel as a cost for sink and a sum over all negative distributions
as a cost for source. The cost for the two neighboring pixels is considered
as a squared difference between the distribution maps for that pixel. The
optimal edge between them is obtained by finding a minimal cut that
maximizes the flow in a graph. We highlight edges with strong borders
between positive and negative distributions by multiplying with
a difference of neighboring pixels in the distribution maps. The resulting
image is finally re-sized and is visualized in the first and the third row of Fig.~\ref{fig:mean-reconstruction-pacman}.

The proposed visualization is applied to the second layer features trained
on PaCMan database. Most features are still representing edges at different orientations,
but some (e.g., 4th and 8th feature) are compositions of edges
in form of corners as is evident from the reconstructed boundaries. The features are visualized on a pruned model,
i.e, we merged any overlapping compositions in each sub-feature to
remove duplicate components, and compositions with weights below a 2\% of a maximal weight are discarded since they do not contribute to
the final score. With pruning we removed or merged approximately
400 out of 1024 Gaussians at the second layer, while reducing
the score by less than 1\%. This pruning process is another benefit of our explicit compositions, which can lead to a network with significantly reduced complexity.

\subsection{Inference speed-up with separable filters}

\begin{figure}
\begin{centering}
\includegraphics[width=0.25\textwidth]{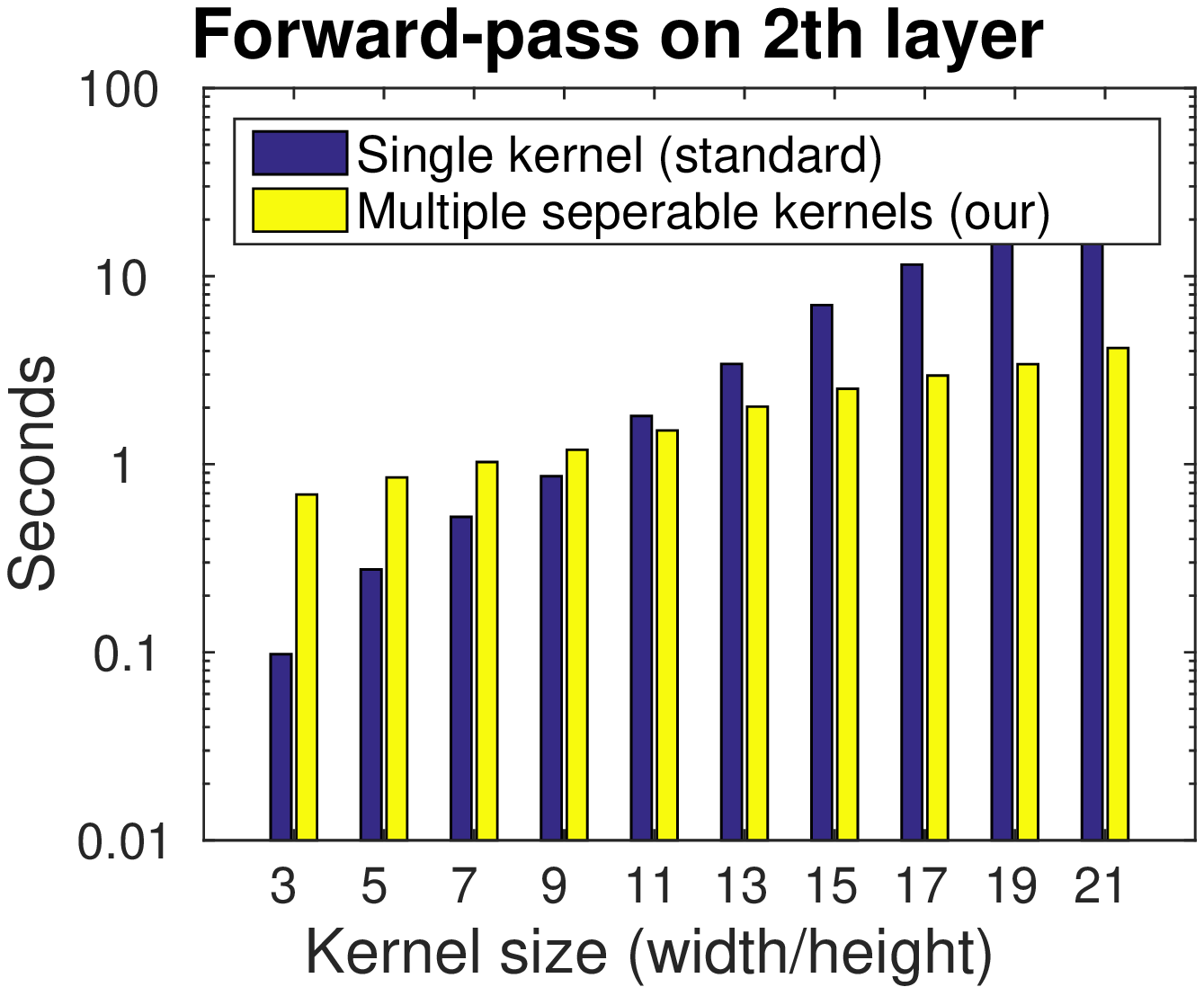}\includegraphics[width=0.25\textwidth]{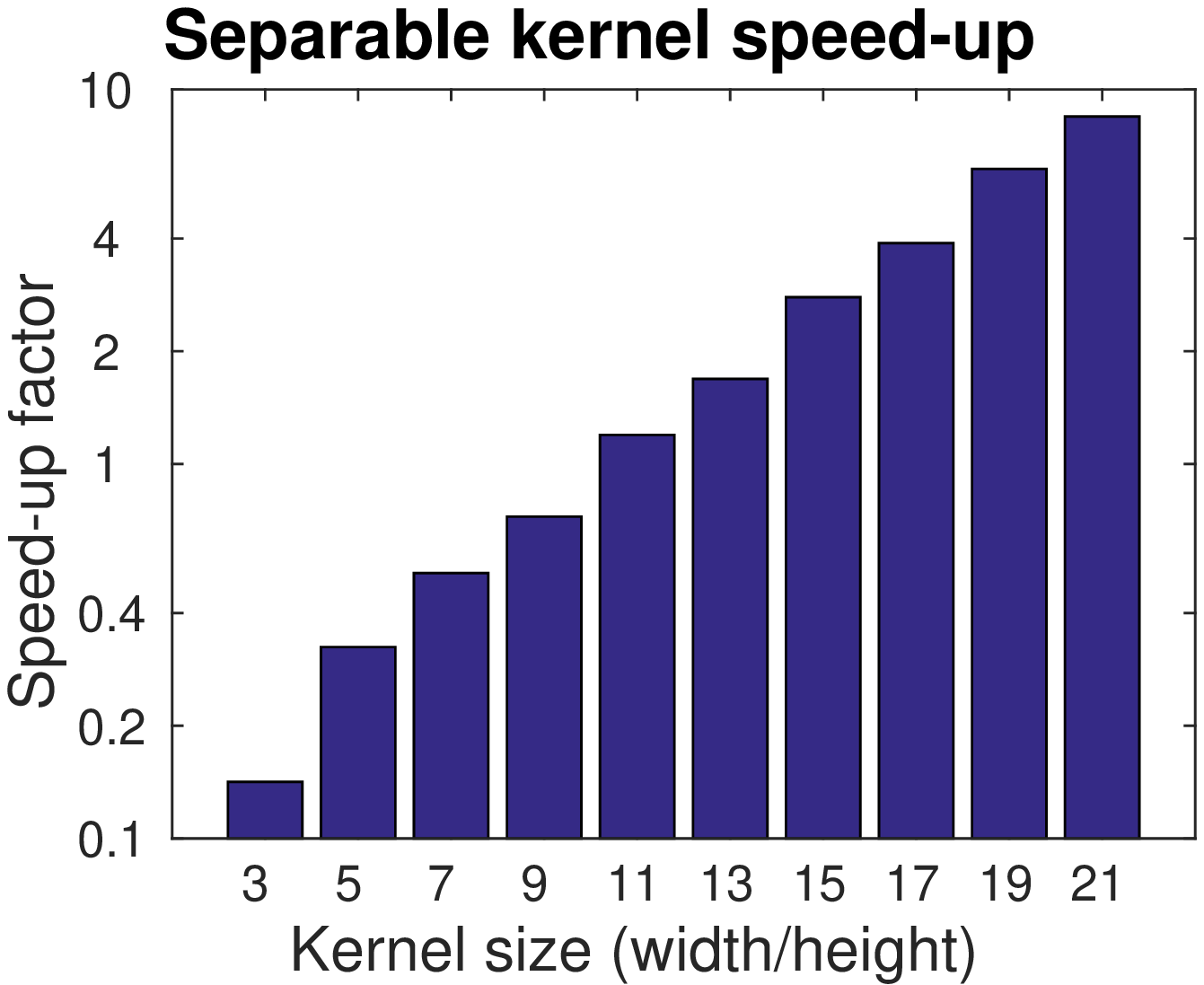}
\par\end{centering}

\caption{\label{fig:speed-up}Forward-pass in second (left) and speed-up factor separable implementation (right). Shown for pruned second layer PaCMan model. Note, the values are shown in the logarithmic scale.}
\vspace{-12pt}
\end{figure}

Another advantage of our deep compositional network is the ability to decompose the units into separate filters by leveraging the separability of Gaussians to speed-up the inference. In this case, the complexity for the forward pass at a single layer is reduced from $O(S\cdot F\cdot W\cdot H\cdot K_{w}\cdot K_{h})$ for the standard CNN to $O(S\cdot F\cdot G\cdot(W\cdot H\cdot(K_{w}+K_{h})+c))$
 for implementation with separable filters, with $S$ sub-features,
$F$ features, $G$ number of Gaussian components per sub-feature,
$W\times H$ feature map size, $K_{w}\times K_{h}$ kernel filter
size and $c$ as an additional overhead in a separable filter implementation.
An important factor in this separable implementation is the number of Gaussian
components $G$, but we can reduce this number with the same
pruning process described in the previous section, where we reduced the
number of components by almost a half. 

Based on time complexities a separable implementation should gain
significant speed-up for $G\cdot(K_{w}+K_{h})<K_{w}\cdot K_{h}$.
We evaluate this separable implementation on a pruned PaCMan model
considering different kernel sizes. Results are depicted in Fig.~\ref{fig:speed-up}.
We achieve a slight speed-up at kernel sizes of $10\times10$ while a significant speed-up of 3-fold or more is achieved with kernel
sizes of $15\times15$ pixels or bigger. 

We use the Caffe implementation with convolution as matrix multiplication
using the CBLAS library and implement separable convolution as multiple
calls to AXPY methods using the same CBLAS library. The demonstrated speed-up
factors are fairly conservative considering the Caffe implementation performs
2D convolution with a fully optimized single call while our implementation
adds some overhead with multiple calls. We evaluated only a CPU implementation and enforced a  single-core process. Since both implementations have a similar level of parallelism it is fair to assume that speed-up can be maintained in multicore CPU or GPU implementations as well.

\section{Conclusion\label{sec:conclusion}}

A new deep compositional network is introduced in this paper. The new network is based on a novel form of an element unit (a filter) that applies a parametric model. We demonstrated that parametrization with Gaussian distributions retains the spatial structure of compositions of features and affords learning by optimizing a well-defined cost function. We derived the necessary equations for back-propagation and embedded our model into a deep neural network framework to evaluate discriminative learning of compositional parts on CIFAR-10 and PaCMan datasets. We showed that having a compositional representation is advantageous for deep networks by presenting a novel visualization and an inference speed-up. We performed visualization of deep network features using a mean reconstruction of parts. Other visualization techniques for deep networks typically rely on approximation with de-convolution~\cite{Zeiler2013} or a complex optimization~\cite{Mahendran2015}, and need to process additional data. In contrast, the compositional representation allowed us to generate a representation of a feature with a fairly simple technique that uses only the model itself and no additional data. A simpler visualization and the 3-fold speed-up of inference speak of the advantages of using the new parametric units in deep networks with a convolutional layered architecture.

In the future we plan to explore other venues opened by combining compositional and convolutional hierarchies, such as, pre-training with co-occurrence statistics, performing  generative and discriminative learning concurrently, and further leveraging inverted indexing of parts for inference.

\let\thefootnote\relax\footnotetext{
\textbf{Acknowledgments.} This work was supported in part by EU IST-600918
project PaCMan and ARRS research project L2-6765. We also acknowledge MoD/Dstl and EPSRC for providing the grant to support the UK academics (Ale\v{s} Leonardis) involvement in a Department of Defense funded MURI project.
}

\bibliographystyle{IEEEtran}
\bibliography{library}

\end{document}